\documentclass[10pt,twocolumn,letterpaper]{article}

\usepackage[pagenumbers]{cvpr} % To force page numbers, e.g. for an arXiv version

\usepackage[dvipsnames]{xcolor}

\usepackage{graphicx}
\usepackage{amsmath}
\usepackage{amssymb}
\usepackage{booktabs}
\usepackage{tabularx}

\graphicspath{{images}}

\definecolor{cvprblue}{rgb}{0.21,0.49,0.74}
\usepackage[pagebackref,breaklinks,colorlinks,citecolor=cvprblue]{hyperref}

\def\paperID{*****} % *** Enter the Paper ID here
\def\confName{CVPR}
\def\confYear{2024}

\title{On the Effect of Image Resolution on Semantic Segmentation}

\author{Ritambhara Singh \qquad Abhishek Jain \qquad Pietro Perona \qquad Shivani Agarwal \qquad Junfeng Yang \\
Department of Computer Science, Duke University\\
{\tt\small rsingh@cs.duke.edu}
}

\begin{document}

\maketitle

\begin{abstract}
High-resolution semantic segmentation requires substantial computational resources. Traditional approaches in the field typically downscale the input images before processing and then upscale the low-resolution outputs back to their original dimensions. While this strategy effectively identifies broad regions, it often misses finer details. In this study, we demonstrate that a streamlined model capable of directly producing high-resolution segmentations can match the performance of more complex systems that generate lower-resolution results. By simplifying the network architecture, we enable the processing of images at their native resolution. Our approach leverages a bottom-up information propagation technique across various scales, which we have empirically shown to enhance segmentation accuracy. We have rigorously tested our method using leading-edge semantic segmentation datasets. Specifically, for the Cityscapes dataset, we further boost accuracy by applying the Noisy Student Training technique.
\end{abstract}

%%%%%%%%% BODY TEXT
\section{Introduction}
Deep convolutional neural networks have set new benchmarks across a range of computer vision applications, including image classification, object detection, semantic segmentation, and human pose estimation, among others. Semantic segmentation, the task of classifying each pixel in an image into a category label, offers a detailed scene understanding by identifying the label, location, and shape of every element within an image. This capability has significant implications for autonomous driving, robotic perception, and other areas.
\par
One of the inherent challenges in semantic segmentation involves balancing the trade-offs between inference resolutions. Certain predictions, especially those requiring attention to minute details like object edges or slender structures, benefit from processing at higher resolutions. This approach allows for finer granularity in the segmentation output. Conversely, the identification of larger structural elements within an image, which necessitates a broader contextual understanding, tends to be more effective at lower resolutions. Here, the network's receptive field can encompass a larger portion of the scene, providing the context needed for accurate large-scale segmentation. This duality highlights the need for adaptable strategies in DCNNs to optimally process and interpret images across different scales and contexts.
\par
Contemporary leading-edge approaches typically downscale input images by factors of 1/2 or 1/4 before processing. This reduction in resolution facilitates the handling of images within the constraints of computational resources—specifically, memory and processing capacity. As a result, lower-resolution processing becomes a practical necessity for complex neural networks. Additionally, feature maps at reduced resolutions tend to be information-rich, contributing to more precise overall network predictions. However, this practice has the significant drawback of omitting fine details.
\par
Some strategies \cite{badrinarayanan2017segnet,ronneberger2015u,khoshsirat2023improving,khoshsirat2023sentence} aim to produce high-resolution outputs by upsampling these lower-resolution feature maps and applying further convolutions. Yet, these methods are prone to overfitting, primarily because the convolutions at higher resolutions are not complemented by analogous operations at lower resolutions, leaving no mechanism within these networks to refine or adjust the ultimate output based on broader contextual information.
\par
More recently developed architectures \cite{yuan2019object, wang2019deep, pohlen2017full} incorporate intricate designs that have shown promising results. Despite their advancements, the complexity of these models presents a significant challenge: it remains impractical to train them directly on high-resolution images due to the extensive computational demands. This limitation underscores the ongoing challenge in neural network design: balancing the need for detailed image processing with the practical constraints of available computational resources.
\par
Numerous approaches in the field leverage network architectures that have demonstrated efficacy in image classification tasks, including variations of ResNet \cite{zhao2017pyramid} and EfficientNet \cite{mohan2020efficientps}. Utilizing networks pre-trained on auxiliary classification tasks offers a significant advantage by initializing a substantial portion of the model's weights, thereby shortening the training duration and frequently leading to better performance than models trained from scratch, especially when the data for the specific application may be scarce.
\par
However, a principal challenge with adopting such pre-trained models is the constraints they impose on the innovation of new methodologies. These limitations are particularly evident in the inability to seamlessly integrate novel network components, such as batch normalization \cite{ioffe2015batch} or innovative activation functions, into the pre-existing architectures. This restriction can stifle the development of unique and potentially more effective approaches by limiting the exploration of the architectural design space.
\par
Recent works \cite{tao2020hierarchical,xie2020self,khoshsirat2023empowering} adopt the Noisy Student Training where a teacher model is trained using the labeled images.
Then, the teacher model is used to generate pseudo labels on unlabeled images.
And finally, a student model is trained to minimize the combined cross entropy loss on both labeled images and unlabeled images.
\par
In this study, we introduce a streamlined network architecture that demonstrates enhanced generalizability and is capable of producing high-resolution segmentation outputs directly. Leveraging a classifier network that has been pre-trained on a comprehensive dataset like ImageNet \cite{russakovsky2015imagenet}, our design benefits from the robust foundational knowledge gained from extensive pre-training. We rigorously evaluate our model across several prestigious semantic segmentation datasets, including Mapillary Vistas \cite{neuhold2017mapillary}, Cityscapes \cite{cordts2016cityscapes}, CamVid \cite{BrostowFC:PRL2008}, COCO \cite{lin2014microsoft}, and PASCAL-VOC2012 \cite{pascal-voc-2012}, to validate its effectiveness.
\par
Our findings demonstrate that a simplified yet strategically designed network, by directly delivering high-resolution segmentations, can attain leading-edge performance metrics. Specifically, within the Cityscapes dataset, we further refine our model's precision by incorporating the Noisy Student Training algorithm, showcasing a significant advancement in accuracy. This approach underscores the potential of minimalist designs in achieving remarkable results in semantic segmentation tasks.

\begin{figure*}
\begin{center}
\includegraphics[width=1.0\linewidth]{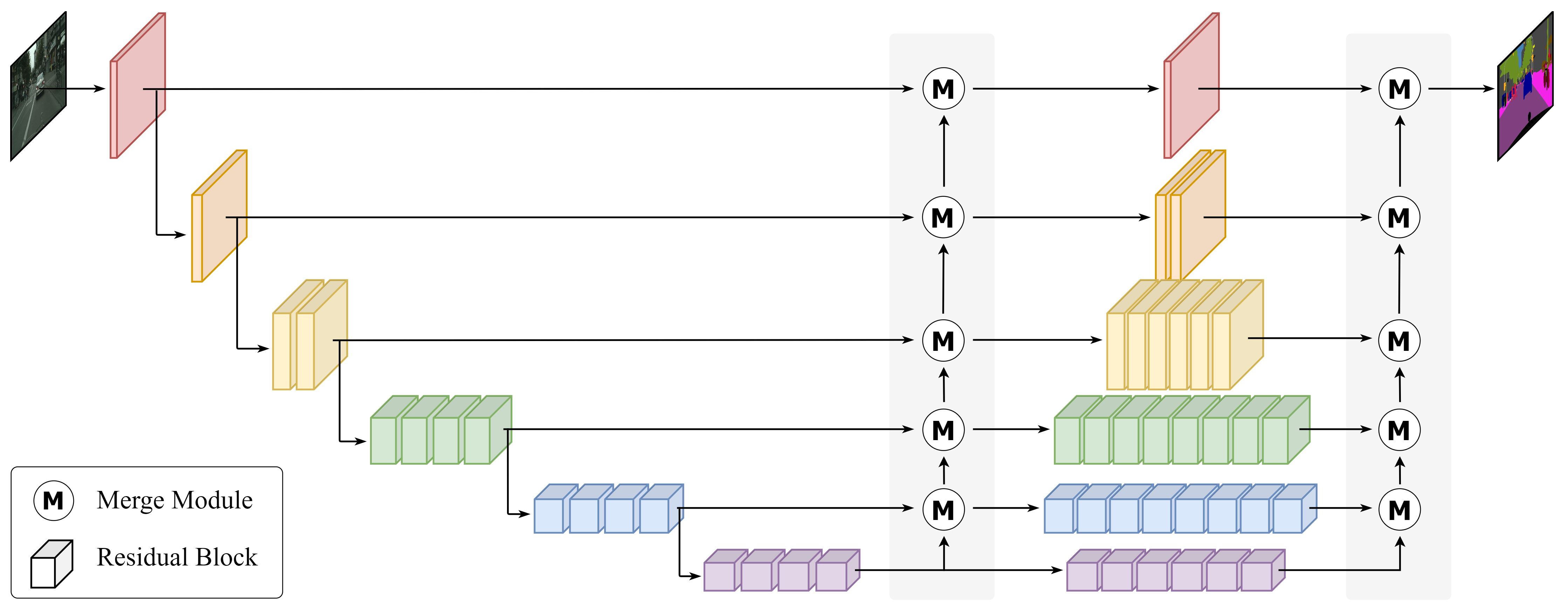}
\end{center}
\caption{The proposed network.
The residual blocks are the same as in \cite{he2016deep}.
The gray areas are the bottom-up propagation stages.}
\label{fig:network}
\end{figure*}

\section{Related Work}
Current state-of-the-art methods for semantic segmentation are based on convolutional neural networks.
These networks have different architectures.
Encoder-decoder or hourglass networks are used in many computer vision tasks like object detection \cite{lin2017feature,khoshsirat2023transformer}, human pose estimation \cite{newell2016stacked,khoshsirat2022semantic}, image-based localization \cite{melekhov2017image,hosseini2022application}, and semantic segmentation \cite{long2015fully,badrinarayanan2017segnet,noh2015learning}. Generally, they are made of an encoder and decoder parts such that, the encoder gradually reduces the feature maps resolution and captures high-level semantic information, and the decoder gradually recovers the low-level details.
Because these networks lose the image details during the encoder path, they are not able to achieve the highest results without using skip connections.
In U-net \cite{ronneberger2015u} by reusing the feature maps from the encoder part of the network (skip connections), low-level image details are recovered.
In U-net's decoder, high-resolution convolutions are not supervised by lower resolution convolutions. so, going deeper by adding convolutional layers does not substantially improve the accuracy.
Spatial pyramid pooling models perform spatial pyramid pooling \cite{lazebnik2006beyond,grauman2005pyramid,maserat201743} at different grid scales or apply several parallel atrous convolution \cite{chen2017deeplab} with different rates. These models include the two famous PSPNet \cite{zhao2017pyramid} and DeepLab \cite{chen2017rethinking}.
DeepLabv3+ \cite{chen2018encoder} adds one skip connection to utilize some of the low-level image details.
High-resolution representation networks \cite{wang2019deep,huang2017multi,fourure2017residual,zhou2015interlinked} try to maintain a high-resolution hidden state from input to output. By doing low-resolution convolutions in parallel streams, high-level features are gained while low-level details are not lost.
Since these networks require a lot of memory, they first downsample the input image to a lower resolution before the main body.
\par
Some approaches \cite{chen2017deeplab,chandra2016fast,khoshsiratembedding} do post-processing, such as conditional random fields, on the network's output to improve the segmentation details, especially around the object boundaries.
These approaches add some processing overhead to training and testing.
Pyramid pooling techniques learn square context regions because pooling and dilation are typically employed in a symmetric fashion.
However, relational context methods build context by attending to the relationship between pixels and are not bound to square regions.
This nature of relational context methods allow context to be built based on image composition.
Such techniques can build more appropriate context for unusual semantic regions, such as scattered regions or a tall thin column.
OCRNet \cite{yuan2019object}, DANet \cite{xue2019danet}, CFNet \cite{zhang2019co} augment the representation for each pixel by aggregating the representations of the contextual pixels, where the context consists of all the pixels.
These works consider the relation (or similarity) between the pixels, which is based on the self-attention scheme \cite{wang2018non, vaswani2017attention}, and perform a weighted aggregation with the similarities as the weights.
These methods are used as an extension to an existing segmentation method.
\par
Self-training has been previously used to improve classification networks \cite{yalniz2019billion}.
In \cite{xie2020self}, self-training with Noisy Student algorithm is used to achieve a new state-of-the-art on ImageNet \cite{russakovsky2015imagenet}.
In Cityscapes, a significant amount of each coarse image is unlabelled due to the coarseness of the labels.
In \cite{tao2020hierarchical}, authors use Noisy Student Training but they generate hard thresholded labels instead of soft labels for the Cityscapes coarse set.
This is because storing soft labels for the Cityscapes coarse set requires around 3.2 TB of storage space \cite{tao2020hierarchical}.
\par
Based on the pros and cons of the aforementioned methods, in this work we design an independent network which processes images at the original, high-resolution input, and directly outputs the high-resolution segmentation with high generalization ability.

\section{Method}
We introduce a dual-component network architecture consisting of an initial classifier unit followed by a dedicated segmentation head.

\subsection{Classifier Network}
The initial segment of our proposed framework functions as a classifier network, designed to produce six feature maps across a spectrum of resolutions, ranging from high to low. Traditional classifiers typically downscale the input resolution early in the process, resulting in the absence of high-resolution feature maps in their outputs. To address this limitation, we have developed a custom network that closely mirrors the architecture of ResNet-34 \cite{he2016deep}, albeit with minor adjustments to better suit our requirements, as illustrated in Figure \ref{fig:network}. Specifically, we enhance the network by integrating additional residual blocks aimed at refining the processing capabilities at the first and second scales.

\begin{figure}[t]
\begin{center}
\includegraphics[width=1.0\linewidth]{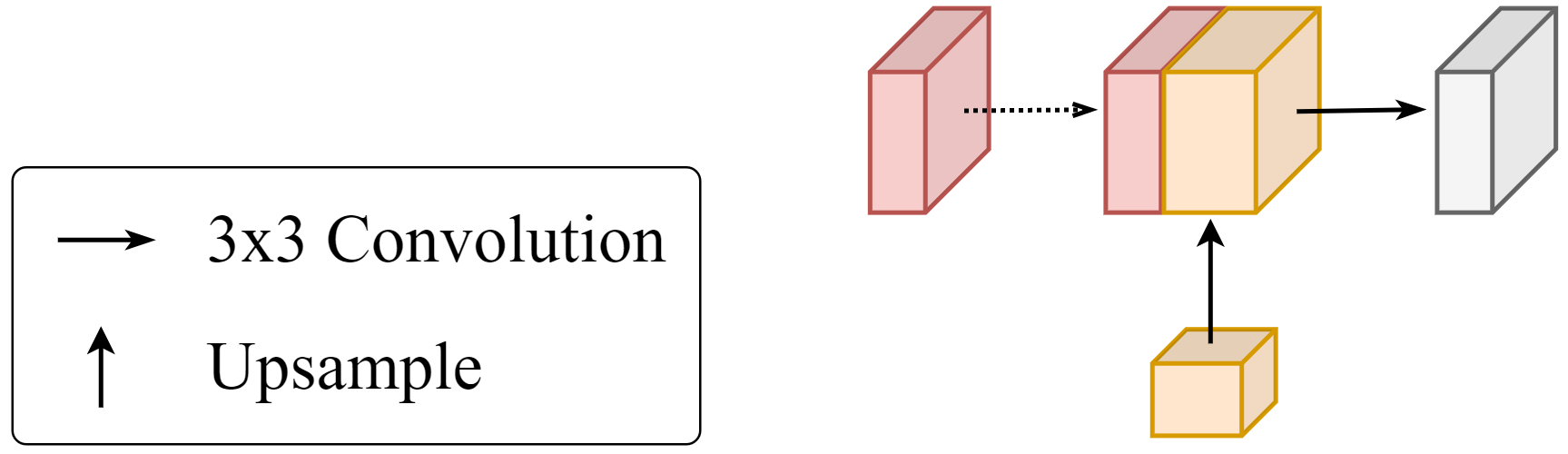}
\end{center}
\caption{The merge module used in the proposed network.
For the upsampling, bilinear interpolation is used to avoid the checkerboard artifact \cite{odena2016deconvolution}.}
\label{fig:merge-module}
\end{figure}

\subsection{Segmentation Head}
The latter portion of our architecture is dedicated to the task of semantic segmentation. Following the classifier network's operation, which yields primary feature maps alongside their associated classes for broadly defined regions, the focus shifts towards refining these maps to generate the definitive segmentation output. Initially, the process involves the upward propagation of information from the coarser, low-resolution feature maps to their higher-resolution counterparts, a technique we refer to as Bottom-Up Propagation. This approach enriches the higher-resolution feature maps with greater contextual depth and an expanded receptive field.
\par
Subsequently, the architecture employs a series of stacked residual blocks \cite{he2016deep} to construct final segmentations across various scales. The culmination of this process involves applying the Bottom-Up Propagation technique once again, this time to amalgamate the multi-scale segmentations into a singular, high-resolution outcome. This methodological framework ensures that the resulting segmentation is both detailed and contextually informed, leveraging the strengths of multi-scale processing to achieve superior accuracy.
\par
{\bf Bottom-Up Propagation.} The process commences by feeding the two feature maps of the lowest resolutions into a Merge Module. Subsequently, the output of this module, along with the next feature map in ascending order of resolution, is input into another Merge Module. This sequential operation continues until it encompasses the feature map with the highest resolution, ensuring a progressive enhancement of detail and context as the resolution increases.
\par
{\bf Merge Module.} Within the Merge Module, the feature map of lower resolution undergoes bilinear interpolation to match the dimensions of its higher-resolution counterpart. Following this resizing, the two feature maps are concatenated to form a unified representation. To streamline the combined feature map for efficient processing, a convolutional layer is then applied to condense the channel count, effectively refining the feature integration. This procedure is graphically depicted in Figure \ref{fig:merge-module}, illustrating the transformation and consolidation steps integral to the Merge Module's function.

\subsection{Noisy Student Training}
Building upon the success of recent studies \cite{tao2020hierarchical, xie2020self} that have demonstrated the effectiveness of Noisy Student Training, we incorporate this technique within our framework for the Cityscapes dataset to enhance both the volume and the quality of the dataset. Initially, we train a teacher model on the labeled images, employing the conventional cross-entropy loss method. Following this, the teacher model is utilized to generate pseudo labels for the images in the coarse set.
\par
In alignment with the approach outlined by \cite{tao2020hierarchical}, we opt for generating hard labels rather than soft labels, addressing the challenge of storage constraints. Subsequently, a student model is trained, aiming to minimize the combined cross-entropy loss derived from both the originally labeled images and those adorned with pseudo labels. Within our experimental setup, data augmentation serves as the source of 'noise' for the student model, and we choose not to iterate the process further. This strategy allows us to leverage the robustness introduced by the Noisy Student Training algorithm, thereby improving the model's performance on semantic segmentation tasks.

\section{Experiments}
Our classifier network undergoes an initial pretraining phase on the ImageNet dataset \cite{russakovsky2015imagenet}, establishing a foundational layer of knowledge that is leveraged across all subsequent experiments. We further refine our model through pretraining on the Mapillary Vistas dataset \cite{neuhold2017mapillary}, followed by dedicated training and evaluation phases on the Cityscapes \cite{cordts2016cityscapes} and CamVid \cite{BrostowFC:PRL2008} datasets to assess its performance across different urban scenes.
\par
Additionally, the model is pretrained on the COCO dataset \cite{lin2014microsoft}, with subsequent training and testing conducted on the PASCAL-VOC2012 dataset \cite{pascal-voc-2012}, allowing us to validate its efficacy in a broad range of visual contexts. Across all these experiments, segmentation accuracy is quantified using the standard mean Intersection over Union (mIoU) metric, providing a consistent measure of the model's ability to accurately delineate and classify various elements within an image.

\begin{table}
\begin{center}
\begin{tabularx}{\columnwidth}{c|c|c|c}
  Scale
  & Output Size
  & Residual Blocks
  & Channels \\
 \hline
 1 & $H \times W$ & 1 & 50 \\
 \hline
 2 & $H/2 \times W/2$ & 1 & 75 \\
 \hline
 3 & $H/4 \times W/4$ & 2 & 125 \\
 \hline
 4 & $H/8 \times W/8$ & 4 & 200 \\
 \hline
 5 & $H/16 \times W/16$ & 4 & 320 \\
 \hline
 6 & $H/32 \times W/32$ & 4 & 450 \\
 \hline
\end{tabularx}
\end{center}
\caption{Our classifier network architecture.
We use the same residual blocks as in \cite{he2016deep}.
Downsampling is performed by scales 2 to 6.}
\label{tab:classifier-network}
\end{table}

\subsection{ImageNet}
Table \ref{tab:classifier-network} presents the configuration of our classifier network in detail. For training image preparation, we employ a data augmentation strategy consistent with the one described by Russakovsky et al. \cite{russakovsky2015imagenet}. Specifically, image resizing is conducted by randomly selecting the length of the shorter side from the range [256, 480], followed by cropping to dimensions of 224$\times$224. Subsequent steps include the application of random horizontal flips and conventional color augmentation techniques to enhance model robustness and generalizability.
\par
The training regimen spans 100 epochs, utilizing batches of 256 images. Optimization is carried out using Stochastic Gradient Descent (SGD) with a weight decay parameter set at 0.0001. The initial learning rate is established at 0.1, with scheduled reductions by a factor of 10 occurring at the 30th, 60th, and 90th epochs, ensuring adaptive learning pace adjustments in response to the evolving training landscape.

\begin{table*}[t]
\begin{center}
\setlength\tabcolsep{3.0pt}
\footnotesize
\begin{tabular}{l|ccccccccccccccccccc|c}
  Method & \rotatebox{90}{Road} & \rotatebox{90}{Sidewalk} & \rotatebox{90}{Building} & \rotatebox{90}{Wall} & \rotatebox{90}{Fence} & \rotatebox{90}{Pole} & \rotatebox{90}{Traffic Light} & \rotatebox{90}{Traffic Sign} & \rotatebox{90}{Vegetation} & \rotatebox{90}{Terrain} & \rotatebox{90}{Sky} & \rotatebox{90}{Person} & \rotatebox{90}{Rider} & \rotatebox{90}{Car} & \rotatebox{90}{Truck} & \rotatebox{90}{Bus} & \rotatebox{90}{Train} & \rotatebox{90}{Motorcycle} & \rotatebox{90}{Bicycle} & \rotatebox{90}{mIoU} \\
 \hline
 ResNeSt200 \cite{zhang2020resnest}          & 98.9 & 88.4 & 94.3 & 66.0 & 66.0 & 72.5 & 78.6 & 82.5 & 94.2 & 72.9 & 96.3 & 88.4 & 74.8 & 96.6 & 77.0 & 92.3 & 90.0 & 73.2 & 79.1 & 83.3 \\
 GALD-Net \cite{li2019global}                & 98.8 & 87.7 & 94.2 & 65.0 & 66.7 & 73.1 & 79.3 & 82.4 & 94.2 & 72.9 & 96.0 & 88.4 & 76.2 & 96.5 & 79.8 & 89.6 & 87.7 & 74.1 & 79.9 & 83.3 \\
 EfficientPS \cite{mohan2020efficientps}     & 98.8 & 88.2 & 94.3 & 67.6 & 67.7 & 73.4 & 80.2 & 83.3 & 94.3 & 74.4 & 96.0 & 88.7 & 75.3 & 96.6 & 83.5 & 94.0 & 91.1 & 73.5 & 79.6 & 84.2 \\
 Panoptic Deeplab \cite{cheng2019panoptic}   & 98.8 & 88.1 & 94.5 & 68.1 & 68.1 & 74.5 & 80.5 & 83.5 & 94.2 & 74.4 & 96.1 & 89.2 & 77.1 & 96.5 & 78.9 & 91.8 & 89.1 & 76.4 & 79.3 & 84.2 \\
 \hline
 Ours                                        & 98.8 & 87.5 & 94.3 & 65.7 & 65.5 & 72.9 & 80.0 & 82.7 & 94.2 & 73.1 & 96.0 & 88.9 & 75.6 & 96.8 & 81.2 & 93.7 & 90.3 & 73.2 & 79.8 & 83.7 \\
 \hline
\end{tabular}
\end{center}
\caption{Results on Cityscapes \cite{cordts2016cityscapes} test set.
All the methods are pretrained on Mapillary Vistas \cite{neuhold2017mapillary}.}
\label{tab:cityscapes-results}
\end{table*}

\subsection{Mapillary Vistas}
The Mapillary Vistas dataset (research edition) \cite{neuhold2017mapillary} serves as a comprehensive resource for street-level image analysis, featuring 25,000 images with dense annotations. These are divided into sets of 18,000 for training, 2,000 for validation, and 5,000 for testing. The dataset encompasses 65 object categories alongside a single void class, with the images presenting a diverse array of aspect ratios and resolutions, extending up to 22 Megapixels.
\par
For training, we employ Stochastic Gradient Descent (SGD) with a weight decay parameter of 0.0001 and a batch size of 10. The learning rate follows a polynomial adjustment policy, characterized by a poly exponent of 0.9 and an initial rate of 0.01, over a total of 500 epochs. Data augmentation techniques are applied to enhance model robustness, including random cropping to dimensions of $768\times768$, scaling within the range of [0.5, 2.0], and random horizontal flipping. Notably, our training process utilizes both the designated training and validation sets to maximize the learning potential from the available data.

\subsection{Cityscapes}
The Cityscapes dataset \cite{cordts2016cityscapes} is a benchmark collection of 5,000 high-definition street images, meticulously annotated at the pixel level for detailed semantic analysis. The dataset is stratified into subsets comprising 2,975 training images, 500 validation images, and 1,525 test images that are finely annotated. Additionally, it includes 20,000 images with coarse annotations, broadening the scope for model training and evaluation. Out of the 30 categorically distinct classes provided, 19 are earmarked for evaluation purposes.
\par
For comprehensive assessment on the test set, we report not only the mean Intersection over Union (mIoU) for class-wise accuracy but also three supplementary scores: IoU category, iIoU class, and iIoU category, offering a nuanced insight into model performance across different segmentation categories.
\par
The training regimen employs Stochastic Gradient Descent (SGD) with a weight decay setting of 0.0001 and a batch size of 10. We adhere to a "polynomial" learning rate adjustment strategy with a poly exponent of 0.9 and an initial learning rate of 0.01, spanning a total of 300 epochs. To enhance the model's adaptability to varied scene compositions, data augmentation techniques such as random cropping to $768\times768$ pixels, scaling within a [0.5, 2.0] range, and random horizontal flipping are applied.
\par
In pursuit of achieving optimal accuracy, our training strategy encompasses not just the fine-annotated training and validation images, but also the extensive set of coarsely annotated images, thereby leveraging the full spectrum of available data within the Cityscapes dataset.
\par
{\bf Noisy Student Training.}
Our methodology employs a stringent labelling approach, where, for any given pixel, the class with the highest prediction probability from the teacher network is designated as the top class. The decision to accept this prediction as the true label is contingent upon surpassing a specified confidence threshold derived from the teacher network's output probability. Only teacher predictions meeting or exceeding this threshold are considered valid labels; pixels failing to meet this criterion are assigned to an "ignore" class, effectively excluding them from contributing to the loss calculation and subsequent training steps. In our implementation, we have set this confidence threshold at 0.9, striking a balance between precision and coverage in the generated pseudo labels.

\subsection{CamVid}
Compared to Cityscapes \cite{cordts2016cityscapes}, CamVid \cite{BrostowFC:PRL2008} is a very smaller dataset focusing on semantic segmentation for driving scenarios.
The original version is composed of 701 annotated images in 32 classes with size 960$\times$720 from five video sequences.
However, most literature only focuses on the protocol proposed in \cite{badrinarayanan2017segnet} which splits the dataset into 367/101/233 images for training, validation, and test in 11 classes.
We follow this protocol for splitting the dataset and train on the training and validation sets to get the highest accuracy on the test set.
We use SGD with a weight decay of 0.0001 and batch size of 16.
We apply the "polynomial" learning rate policy with a poly exponent of 0.9 and initial learning rate of 0.01, and train for 300 epochs.
The data are augmented by random cropping ($768\times768$), random scaling in the range of [0.5, 2.0], and random horizontal flipping.

\subsection{COCO}
The Microsoft COCO \cite{lin2014microsoft} dataset contains 118k/5k images for training and validation in 80 object categories.
The images have varying aspect ratios and sizes.
We use SGD with a weight decay of 0.0001 and batch size of 12.
We apply the "polynomial" learning rate policy with a poly exponent of 0.9 and initial learning rate of 0.01, and train for 90 epochs.
The data are augmented by random cropping ($480\times480$), random scaling in the range of [0.5, 2.0], and random horizontal flipping.

\subsection{PASCAL-VOC2012}
The PASCAL-VOC2012 \cite{pascal-voc-2012} segmentation dataset contains 20 object categories and one background class.
The dataset has 1,465 training, 1,450 validation, and 1,456 test images.
We augment the dataset by the extra annotations provided by \cite{hariharan2011semantic}, resulting in 10,582 training images.
We use SGD with a weight decay of 0.0001 and batch size of 12.
We apply the "polynomial" learning rate policy with a poly exponent of 0.9 and initial learning rate of 0.01, and train for 300 epochs.
The data are augmented by random cropping ($480\times480$), random scaling in the range of [0.5, 2.0], and random horizontal flipping.

\section{Results}
Our method has been evaluated across multiple datasets, with the outcomes benchmarked against leading state-of-the-art techniques.

\subsection{Cityscapes}
The performance of our approach on the Cityscapes \cite{cordts2016cityscapes} test set is detailed in Table \ref{tab:cityscapes-results}. Each method under comparison has undergone pretraining on the Mapillary Vistas dataset \cite{neuhold2017mapillary} to ensure a consistent basis for evaluation. Our training regimen encompasses the comprehensive use of Cityscapes' training, validation, and coarse annotation sets to maximize the learning potential from the available data.

\section{Conclusion}
In this study, we introduced a simplified network architecture that directly produces high-resolution semantic segmentations, challenging the conventional methodology that relies on downscaling and upscaling processes. Our findings demonstrate that this streamlined approach not only matches but, in certain instances, surpasses the performance of more complex, lower-resolution systems. Key to our success is the implementation of a bottom-up information propagation strategy, which effectively enhances segmentation accuracy by ensuring that higher-resolution feature maps are informed by contextually rich, lower-resolution counterparts.
\par
Extensive experimentation across various leading semantic segmentation datasets has validated the efficacy of our model. Notably, our application of the Noisy Student Training technique on the Cityscapes dataset signifies a notable advancement in segmentation accuracy, showcasing the potential of leveraging semi-supervised learning methods within high-resolution segmentation tasks.
\par
The implications of our research are twofold. Firstly, it affirms that high-resolution segmentation can be achieved without the computational penalties traditionally associated with processing at native image resolutions. Secondly, it underscores the versatility and adaptability of simplified network structures in handling complex visual tasks, opening avenues for further exploration in efficient network design and training methodologies.
\par
Future work will focus on refining the bottom-up propagation mechanism to further enhance detail capture and exploring the integration of additional semi-supervised and unsupervised learning techniques to expand the model's applicability and robustness across diverse and challenging segmentation scenarios.

{
    \small
    \bibliographystyle{ieeenat_fullname}
    \bibliography{main}
}

\end{document}